%% file: main.tex
\documentclass[conference]{IEEEtran}
\IEEEoverridecommandlockouts
\usepackage[letterpaper, left=0.75in, right=0.75in, bottom=0.75in, top=.75in]{geometry}
\usepackage{cite}
\usepackage{amsmath,amssymb,amsfonts}
\usepackage{todonotes}
\usepackage{graphicx}
\usepackage{textcomp}
\usepackage{xcolor}
\usepackage{algorithm}
\usepackage{algpseudocode}
\usepackage{nicematrix}
\usepackage{caption}
\def\BibTeX{{\rm B\kern-.05em{\sc i\kern-.025em b}\kern-.08em
    T\kern-.1667em\lower.7ex\hbox{E}\kern-.125emX}}

\newcommand{\ForParallel}[1]{\For{\textbf{ parallel #1}}}
\newcommand{\EndForParallel}{\EndFor{\textbf{ parallel}}}

\newcommand{\state}{\mathbf{s}}
\newcommand{\ctrl}{\mathbf{u}}
\newcommand{\pos}{\mathbf{p}}
\newcommand{\vel}{\mathbf{v}}
\newcommand{\Ubox}{\mathbf{U_{box}}}

\newtheorem{theorem}{Theorem}
\newtheorem{lemma}[theorem]{Lemma}

\newcommand{\Reals}{\mathbb{R}}
\newcommand{\ourmethod}{FiReFly}
    
\begin{document}

\title{
\vspace*{.35cm}
\ourmethod: Fair Distributed Receding Horizon Planning for Multiple UAVs
\thanks{This work was partially supported by the FAA ASSURE Center of Excellence under projects A51 and A58, NSF Award 2118179 and by Toyota Research Institute of North America.}
}

\author{\IEEEauthorblockN{Nicole Fronda}
\IEEEauthorblockA{
\textit{Oregon State University}\\
frondan@oregonstate.edu}
\and
\IEEEauthorblockN{Bardh Hoxha}
\IEEEauthorblockA{
\textit{Toyota Research Institute of North America}\\
bardh.hoxha@toyota.com}
\and
\IEEEauthorblockN{Houssam Abbas}
\IEEEauthorblockA{
\textit{Oregon State University}\\
abbasho@oregonstate.edu}
}

\maketitle

\begin{abstract}
We propose injecting notions of fairness into multi-robot motion planning.
When robots have competing interests, it is important to optimize for some kind of fairness in their usage of resources. In this work, we explore how the robots' energy expenditures might be fairly distributed among them, while maintaining mission success. We formulate a distributed fair motion planner and integrate it with safe controllers in a algorithm called \ourmethod. For simulated reach-avoid missions, \ourmethod~produces fairer trajectories and improves mission success rates over a non-fair planner.
We find that real-time performance is achievable up to 15 UAVs, and that scaling up to 50 UAVs is possible with trade-offs between runtime and fairness improvements.
\end{abstract}


\section{Introduction} \label{sec:intro}


Urban Air Mobility (UAM) envisions an ecosystem in which multiple autonomous vehicles -- each having various objectives and operated by different stakeholders -- share the same space. 
Current motion planning approaches for multiple robots focus on mission success and safety. These approaches may produce plans that get a few robots to their destination quickly, but have others spend excessive energy loitering, or navigating unnecessarily longer paths before reaching their target. This is clearly an unfair outcome for all but the single operator whose robot achieved greater energy efficiency. Fairness is not a primary concern of existing approaches, so we cannot expect fair motion plans to readily emerge from them. A fairer plan would distribute the energy expenditure more evenly between the robots, while still fulfilling the mission objectives, and without compromising other mission constraints. Fairness can be defined in terms of different resources. In this work, we look at fairness in terms of energy. 

There are two reasons at least for considering energy fairness in UAM. 
It would ensure UAVs carrying human passengers do not expend more energy than necessary, preventing the vehicle from dropping to safety-critical fuel or battery levels. Additionally, systematically integrating fairness notions into UAV planning promises equitable experiences to all operators, some of whom would otherwise face increased operational costs from energy-inefficient paths. Fair and equitable operations 
promote greater participation and diversity in UAM, which furthers innovation of UAV technologies.

\begin{figure}[htp]
\centering
{\includegraphics[width=.8\linewidth]{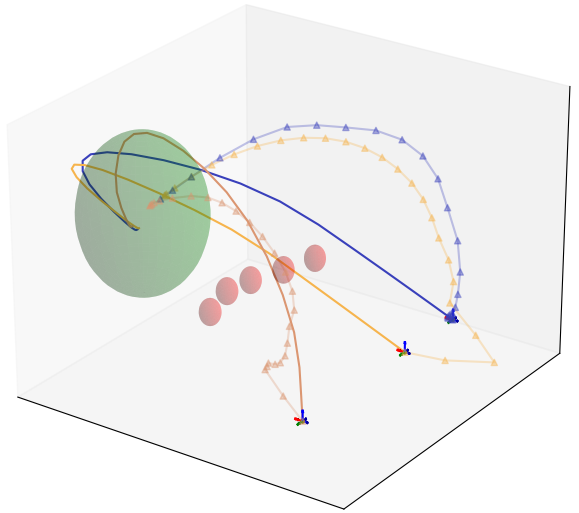}}
\caption{Trajectories for 3 UAVs traveling to a goal (green sphere). The trajectories adjusted by a safe controller (lines with triangle markers) using a CBF cause some UAVs to spend an unfair amount of energy relative to the others. \ourmethod~produces trajectories (solid lines) that are safe \textit{and} fair.}
\label{fig:traj}
\end{figure}

In this paper, we present a distributed motion planning algorithm that explicitly optimizes for fairness in normalized energy consumption across multiple UAVs in the same airspace. A distributed algorithm is necessary in settings where a centralized planner is unavailable or has limited capacity, as in the case of un-towered air sectors, or busy airspaces that must prioritize traditional aircraft. We integrate this fair distributed algorithm into a framework that guarantees safe control, relying on Control Barrier Functions (CBFs) in a receding horizon fashion to ensure collision avoidance. We call this solution \textbf{\ourmethod}, for \textbf{F}air Distributed \textbf{Re}ceding Horizon Planning for \textbf{Fly}ing robots.
Example safe \emph{and} fair \ourmethod~trajectories are shown in Fig. \ref{fig:traj}.

\textbf{Related Work}
An overview of general UAV path planning techniques is presented in the surveys \cite{multiuav_opt_survey,drones9040263}, but these approaches do not consider any notions of fairness.


FairFly \cite{kurtz20fairfly} introduced an offline (before flight) optimization method for fair UAV motion planning with fairness in terms of flight time. In this paper, we present an online optimization for fairness in terms of energy consumption, which is not always proportional to flight time since it depends on flight characteristics. Safety in \cite{kurtz20fairfly} is guaranteed offline via an expensive global optimization of robustness from \cite{pant_2018}. We guarantee safety online (during flight) using Control Barrier Functions (CBFs), which have been shown to be capable of real-time control \cite{ames_2017}.



The work in \cite{brandao_fair_nav_planning} presents different formalizations for fairness in robot planning in terms of locations visited or groups served, which are mathematically different from our defined notions for fairness in energy. Fair energy consumption for multi-UAV teams is explored in \cite{fair_positioning}, but for multi-target positioning, and also uses a different formulation for fairness.

Other energy-aware multi-UAV planners include those in \cite{zhao2021_energyefficientcoverage,buyukkocak_energyaware}, but these focus on total energy efficiency of a UAV team, and not fairness. These notions are not reducible to each other; plans that have low total energy can be still be unfair to some agents, and vice versa. Our approach looks at the energy consumption of individual UAVs, and attempts to distribute it between them equitably.

\textbf{Contributions}
\label{sec:contributions}
This paper addresses the question of fair motion planning for UAVs in the same airspace.
\begin{enumerate}
    \item We define four new notions of fairness which seek to balance normalized energy expenditures
    \item We provide a computational formulation and distributed solution to the fair planning problem in reach-avoid missions, and 
    \item An algorithmic framework called \textbf{\ourmethod} that minimally modifies fairly generated trajectories for safety. We evaluate \ourmethod~in simulation and compare it to a non-fair baseline and state-of-the-art fair planner.
\end{enumerate}


\section{Problem Statement}\label{sec:prelim}
\input{sections/problem_statement}

\section{Fair Motion Planning}\label{sec:fair_planner}
\input{sections/fair_planner}

\section{FiReFly: Fairness with Safety Guarantees}\label{sec:firefly_framework}
\input{sections/firefly_framework}

\section{Experiments}\label{sec:experiments}
\input{sections/experiments}

\section{Conclusion}\label{sec:conclusion}
We address fair motion planning for multiple UAVs by injecting notions of fairness into the planning optimization. 
We formalize four fairness notions to balance normalized energy and power expenditure across UAVs. 
We provide a distributed solution for the fair motion problem as well as an algorithm, FiReFly, which ensures safety. We evaluate FiReFly in reach-avoid missions with increasing number of obstacles and UAVs. 
We found that explicitly including fairness in the problem formulation can produce fairer trajectories, and even improve mission success, over an non-fair baseline. Future work includes incorporating fairness in more sophisticated control architectures using quadrotor dynamics, and in teams with heterogeneous dynamics to reflect the real-world diversity of UAV models and their varying capabilities and energy profiles. Other future work includes analysis of how modeling errors over time affect fairness performance.



\bibliographystyle{IEEEtran}
\bibliography{references}

\section{Appendix}
\input{sections/appendix}

\end{document}

%% file: sections/problem_statement.tex
\subsection{System Model}
\label{sec:system model}
We consider a team of $N \geq 2$ robot agents modeled with discrete-time 
double integrator dynamics with states $\state_k[t]$ and control inputs $\ctrl_k[t]$, for $1\leq k\leq N$, with sample time $d_t$
\begin{equation}
  \state_k[t+1] = \mathbf{A}_k\state_k[t] + \mathbf{B}_k\ctrl_k[t], \nonumber
\end{equation}

\begin{equation}
    \mathbf{A_k} = \begin{bNiceArray}{cccccc}[small, columns-width = 2mm ]
        \Block{3-3}{\mathbf{I}} & & \Block{3-3}{d_t\mathbf{I}} \\
        & & & & & \\
        & & & & & \\
         \Block{3-3}{\mathbf{0}} & & \Block{3-3}{\mathbf{I}} \\
        & & & & & \\
        & & & & &
    \end{bNiceArray},~
    \mathbf{B_k} = \begin{bNiceArray}{ccc}[small, columns-width = 3mm ]
    \Block{3-3}{\frac{d_t^2}{2}\mathbf{I}} & & \\
    & & \\
    & & \\
    \Block{3-3}{d_t\mathbf{I}} & & \\
    & & \\
    & & \\
    \end{bNiceArray} \nonumber
\end{equation}

The state vector $\state_k[t]$ models position and velocity in the x,y,z directions: 
$\state_k[t]=(p_k^x[t], p_k^y[t], p_k^z[t], v_k^x[t], v_k^y[t], v_k^z[t])^T$. The vector $\pos_k[t]$ gives just the position of robot $k$ at time $t$: $\pos_k[t] = (p_k^x[t], p_k^y[t], p_k^z[t])^T$. The vector $\vel_k[t]$ gives just the velocity of robot $k$ at time $t$:  $\vel_k[t] = (v_k^x[t], v_k^y[t], v_k^z[t])^T$.
The input vector $\ctrl_k[t]$ represents acceleration in the x,y,z directions, $\ctrl_k[t]=(a_k^x[t], a_k^y[t], a_k^z[t])^T$.  
This model gives us a clear mapping between control effort and motion, which allows us to analyze the fundamental trade-offs between fair energy distribution and mission success.

The inputs are subject to a convex constraint $\ctrl_k[t] \in \Ubox \subset \Reals^m$ for all $t$.
Let $\ctrl[t]$ be the concatenation of all robots' control inputs at time $t$: $\ctrl[t] = (\ctrl_1[t], \ctrl_2[t], \ldots, \ctrl_N[t]) \in \Reals^{N \times m}$. Without the time index, we use $\ctrl_k$ to refer to the full sequence of control inputs  for robot $k$.
Let $\ctrl$ be the concatenation of all robot control sequences: $\ctrl = (\ctrl_1,\ldots,\ctrl_N)$.
We also define a \textit{prefix} of control inputs $\ctrl^{\leq t}$ that is the concatenation of the history of control inputs up to time $t$: $\ctrl^{\leq t} = (\ctrl[0], \ctrl[1], \ldots \ctrl[t])^T$. For a prefix $\ctrl^{\leq t}$ and $t' \leq t$, let $\ctrl^{\leq t}[t']$ refer to the historical control inputs executed at time step $t'$.
Let $\state[t]$ be the full system state at time $t$: $\state[t] = (\state_1[t], \state_2[t], \ldots, \state_N[t])^T$.


\subsection{Reach-Avoid Mission}

Each robot has a starting position $\pos_k[0] \in \Reals^3$ and a spherical goal set $G_k \subset \Reals^3$ centered at $c_{G_k}$ with radius $r_{G_k}$. Robots must maintain a safe distance $d_s$ between each other or risk collision.
The mission space includes $O$ static obstacles modeled as spheres with centers $c_o \in \Reals^3$ and radii $r_o$, $1\leq o\leq O$. 
Other static obstacles that are non-spherical can be over-approximated by a union of spheres. 
Robot $k$'s objective is to \textit{reach} $G_k$ in at most $H_k$ time-steps and \textit{avoid} collisions with obstacles and each other. 

\subsection{Fairness}
In this paper, fairness is a property $\ctrl$ and 
fairness notions are modeled as functions $f: \Reals^{ m\sum_{k\leq N} H_k} \rightarrow \Reals$.
We explore different notions of fairness in Section \ref{sec:fairness}.

\subsection{Problem Statement}
Given $N$ robots and a mission space composed of goal and obstacle sets, compute $N$ control sequences in $\Ubox$ such that each robot $k$ reaches its goal set $G_k$ within $H_k$ time steps without collisions, and such that fairness $f$ is maximized. 

\subsection{Solution Overview} \label{sec:solution_overview}
Solving for fair and safe control inputs simultaneously in a single optimization would require satisfaction of multiple non-linear constraints, which is computationally prohibitive. We instead deal with satisfying linear constraints by handling the fairness and safety problems separately. We take an iterative, receding horizon approach to solving the problem that proceeds in two steps.
1) Optimize for fairness: at time $t$, compute $N$ control input sequences (one per robot) such that each robot reaches its goal set in at most $H_k-t$ time steps, and such that fairness $f$ \textit{of the entire set of trajectories} is maximized,  
and 2) Adjust for safety: given these fair input sequences, compute $N$ \textit{one-step safe inputs} for time $t$ such that progress to the goal is maintained, and the difference between these safe inputs and the first inputs from the fair sequences is minimized.
The idea is that the robots execute safe control inputs at every time step and, because the process is iterative, reference trajectories will always be updated to be as fair as possible given the history of inputs.

%% file: sections/fair_planner.tex
We introduce four fairness notions and use them to formulate the optimization problem for computing fair trajectories.

\subsection{Notions of Control Fairness}\label{sec:fairness}
The first two fairness notions are adapted from \cite{kurtz20fairfly}.
Whereas that work applied them to flight time as the resource, we apply them to energy consumption.

\subsubsection{Energy Variance} 
The first notion strives to ensure that each agent consumes a similar amount of energy to complete its mission.
Because different missions can require different amounts of energy to complete it makes little sense to require that all robots consume the same amount of energy. 
Therefore we use normalized energies. 
Let $\ctrl^{solo}_k$ be the inputs of robot $k$ if it is the single robot in the environment. The energy $\underline{e}_k$ that robot $k$ consumes in this case is: $\underline{e}_k = \sum_t\ctrl^{solo}_k[t]^2$.
Then, we define the normalized energy of robot $k$ as 
 $$e_k := \|\ctrl_k\|^2/\underline{e}_k=\frac{1}{\underline{e}_k}\sum_t\ctrl_k[t]^2$$
Normalized energy is always greater than equal to 1, and represents the cost increase due to the presence of others in the environment. Let $\mathbf{e} = (e_1,\ldots, e_N)^T$ be the concatenation of all robots' normalized energies. 
Our first fairness notion is the variance between the normalized energies:
\begin{equation} \label{eq:f1}
f_1(\ctrl) := Var(\mathbf{e}) = \frac{1}{N}\sum_{k}(e_k- N^{-1}\mathbf{1}^T\mathbf{e})^2   
\end{equation}
Perfect fairness is achieved when $f_1(\ctrl)=0$, i.e. whenever all robots consume the exact same normalized energy. 

Function $f_1$ alone might actually cause problematic behavior: minimizing $f_1$ forces all robots to consume the same normalized energy, even if there is a solution in which robot 1, say, consumes less energy without this increasing the others' consumption.
From robot 1's perspective, this is unfair. 
We therefore introduce the following notion, where $\beta>0$ is a hyperparameter and $Q$ is a positive definite matrix.
\begin{equation}
\label{eq:f2}
    f_2(\ctrl) = f_1(\ctrl) + \beta \ctrl^TQ\ctrl
\end{equation}
We refer to $\beta \ctrl^TQ\ctrl$ as an additional \textit{energy term}. Function $f_2$ promotes less control energy, as long as this doesn't unduly impact fairness according to $f_1$.  

\subsubsection{Surge Variance}
Large changes of energy consumed from moment to moment (in other words, power consumption surges) are generally undesirable as they strain the system and jostle payload.
So a third notion of fairness we experiment with seeks to distribute such surges evenly across the robots, once they exceed a threshold. 
Let $M$ be the acceptable surge threshold, i.e., we consider that there is a surge if $|e_k[t] - e_k[t-1]| >> M $.
The total energy surge $z_k$ for a robot $k$ is computed as:
$$z_k = \sum_{t=1}^{H_k}{(|e_k[t] - e_k[t-1]| - M)} $$
Let $\mathbf{z} = (z_1,\ldots, z_N)^T$ be the concatenation of all robots' total energy surge. The resulting fairness notion is the variance of $\mathbf{z}$:
\begin{equation} \label{eq:f3}
f_3(\ctrl) = Var(\mathbf{z}) = \frac{1}{N}\sum_{k}(z_k - N^{-1}\mathbf{1}^T\mathbf{z})^2  
\end{equation}

As before, we also define a notion that combines $f_3$ with the energy term to reduce overall control energy.

\begin{equation} \label{eq:f4}
f_4(\ctrl) = f_3(\ctrl) + \beta \ctrl^TQ\ctrl
\end{equation}

\subsection{Fair Motion Planner}

The fair control problem at time $t\geq 0$ can be formulated as the following optimization problem. With $f$ being one of the functions above:
\begin{subequations}
\label{eq:central_fair}
\begin{align}
\ctrl^{fair} & = \text{arg}\min_{\ctrl} ~ f(\ctrl) \label{eq: central_fair_obj}\\
\text{s.t.} &~ \state_k[t'+1] = \mathbf{A}_k\state_k[t']+\mathbf{B}_k\ctrl_k[t'] ~\forall~ k, t' \label{eq: central_fair_dyn}\\
& ~ \ctrl_k \in \Ubox~\forall~k \label{eq: central_fair_box} \\
& ~ \pos_k[H_k] \in G_k~\forall~k \label{eq: central_fair_goal} \\
& ~ \ctrl[t'] = \ctrl^{\leq t-1}[t'] ~\text{for}~ 0 \leq t' \leq t-1 \label{eq: central_fair_prefix}
\end{align}
\end{subequations}

The objective \eqref{eq: central_fair_obj} maximizes fairness by minimizing function $f$. This is subject to the robots' dynamics \eqref{eq: central_fair_dyn}, the input constraints \eqref{eq: central_fair_box}, and the requirement for each robot to reach their goal by $t=H_k$ \eqref{eq: central_fair_goal}. Constraint \eqref{eq: central_fair_prefix} fixes all the previously executed inputs. Note that non-collision constraints are not in this problem. How these are handled is discussed in the next section.

To solve Problem \ref{eq:central_fair} in a distributed manner, we use a modified version of the algorithm developed in \cite{pant_2022} which is based on \cite{razaviyan14dstr}. The process is iterative, starting from some initial control input $\ctrl^{(0)}$.
At iteration $r$, each robot $k$ solves a \textit{local problem} $P^{(r)}_k$ over some local variables to produce a new estimate $\ctrl_k^{(r)}$. (Problem $P_k^{(r)}$ is given in the next paragraph).
The previous iterates $\ctrl^{(r-1)}_{j\neq k}$ of the other robots appear in $P^{(r)}_k$ as constants.
After generating a solution, robots exchange their local solutions $\ctrl_k^{(r)}$, and the process repeats: robot $k$ re-solves $P^{(r+1)}_k$ with the received values from the other robots set as constants. 
This continues until all local solutions converge, or some maximum number of iterations $R$ is reached. The pseudo-code is given in Algorithm~\ref{alg:distributed_opt}.

The search variable for the local problem, given below, is $\varepsilon_k^{(r)} \in \Reals^m$.
This is the descent vector that robot $k$ applies to the most recent iterate: $\ctrl_k^{(r)}=\ctrl_k^{(r-1)} + \gamma^{(r)} \varepsilon_k^{(r)}$, where $\gamma^{(r)}$ is an adaptive step size which we set to gradually decrease with each iteration from 1 to 0.1. The local problem $P_k^{(r)}$ at iteration $r\geq 1$ for agent $k$ is:

\begin{subequations}
\label{eq:dist_fair}
\begin{align}
\min_{\varepsilon_k^{(r)}} & -\varepsilon_k^{(r)T} \nabla_{\ctrl_k}~ f(\ctrl^{(r-1)}) + \kappa \|\varepsilon_k^{(r)}\|^2 \label{eq: dist_fair_obj} \\
\text{s.t.} &~ \state_k[t'+1] = \mathbf{A}_k\state_k[t']+\mathbf{B}_k (\ctrl_k^{(r-1)} + \varepsilon_k^{(r)})[t'] \label{eq:dist_fair_dyn} \\
&~ \ctrl_k^{(r-1)} + \varepsilon_k^{(r)} \in \mathbf{U_{box}} \label{eq:dist_fair_ubox} \\ 
&~ \varepsilon_{min} \leq \varepsilon_k^{(r)} \leq \varepsilon_{max} \label{eq:dist_fair_eps_bounds} \\
&~ \pos_k[H] \in G_k \label{eq:dist_fair_reach} \\
& ~ \ctrl[t'] = \ctrl^{\leq t}[t'] ~\text{for}~ 0 \leq t' \leq t-1 \label{eq: dist_fair_prefix}
\end{align}
\end{subequations}

The local objective is minimized when the local descent is aligned with the gradient of $f$ (evaluated at the most recent $\ctrl^{(r-1)}$).
Note that the gradient term is a constant within the local problem.
The second term $\kappa\|\ldots\|^2$ encourages small changes in $\varepsilon_k$ for each iteration.
We also add a box constraint for $\varepsilon_k$ with the bounds $\varepsilon_{min}$ and $\varepsilon_{max}$ to ensure feasibility of the global solution. Otherwise, the change to a robot's inputs may be too large, and it is not guaranteed this new $\varepsilon_k$ will not cause another robot's problem to be infeasible at the following iteration. 
We show in the Appendix that the local problem \eqref{eq:dist_fair} converges to a stationary point of $f$.

\begin{algorithm}
\caption{Distributed Fair Motion Planning}
\label{alg:distributed_opt}
\begin{algorithmic}[1]
\State{Input $\ctrl^{(0)}$, $\ctrl^{\sim t}$, $\state[0]$, and $\mathbf{A}_k, \mathbf{B}_k, H_k ~1 \leq k \leq N$, $\eta>0$}
\State{Initialize $\varepsilon_k^{(0)} = 0 ~\forall k~ \in 1,\ldots,N$}
\For{$r = 1, \ldots, R$}
        \ForParallel{$k \in 1, \ldots N$}
            \State{Robot $k$ solves local problem $P_k^{(r)}$}
        \EndForParallel
    \State{Update team inputs: $\ctrl_k^{(r)}=\ctrl_k^{(r-1)}+ \gamma^{(r)} \varepsilon_k^{(r)}, ~\forall k~$}
    \If{$\|\ctrl^{(r)} - \ctrl^{(r-1)}\| \leq \eta$} \Comment{Inputs converge}
        \State{\textbf{break}}
    \EndIf
\EndFor
\State \Return $\ctrl$
\end{algorithmic}
\end{algorithm}

%% file: sections/firefly_framework.tex
We now introduce the FiReFly algorithm for minimally adjusting the generated fair control inputs to guarantee safety. 

First, we define a CBF for static obstacle avoidance following the approach in \cite{magnus_2017_NBF}

\begin{eqnarray} \label{eqn: nbf_obstacle}
h^o(\state[t]) = \min_{k, o} \|\pos_k[t] - c_o\|^2 - r_o^2  \nonumber
\end{eqnarray}

\noindent where $h^o(\state[t]) \geq 0$ implies no robot collided with obstacle $o$. 

We also define a CBF for safe separation of pairs of robots

\begin{eqnarray} \label{eqn: nbf_separation}
h^c(\state[t]) = \min_{k,j} \|\pos_j[t] - \pos_k[t] \|^2 - d_s^2 \nonumber
\end{eqnarray}

\noindent where $h^c(\state[t]) \geq 0$ implies no collision between robots. 

A single CBF combines $h^o$ and $h^c$:

\begin{eqnarray} \label{eqn: nbf_full}
h(\state[t]) = \min( h^o(\state[t]), h^c(\state[t]) )
\end{eqnarray}

To ensure each robot eventually reaches their goal, we define the Control Lyapunov Function (CLF)

\begin{eqnarray} \label{eqn: clf}
V(\state_k[t]) = \max_{k} \|\state_k[t] - c_g \|^2 - r_g^2
\end{eqnarray}

\noindent where $V(\state_k[t]) \leq 0$ implies all robots are within their goal region $G_k$.

At each $t$, given the sequence of fair inputs $\ctrl^{fair}$, computed by Algorithm~\ref{alg:distributed_opt}, find a safe control input $\ctrl^{safe}[t]$ by solving

\begin{subequations}
\label{eq:central_safe}
\begin{align}
    \ctrl^{safe}[t] &= \text{arg}\min_{\ctrl[t], \delta} \| \ctrl[t] - \ctrl^{fair}[t] \|^2 + \delta^2 \label{eqn: online_problem_obj} \\
    \text{s.t.} &~ \nabla h(\state[t])^T(\mathbf{A}\state[t] + \mathbf{B}\ctrl[t]) + \alpha h(\state[t])\ \geq 0 \label{eqn: online_problem_cbf} \\
    & \nabla V(\state[t])^T(\mathbf{A}\state[t] + \mathbf{B}\ctrl[t]) + \lambda V(\state[t])\ \leq \delta \label{eqn: online_problem_clf} \\
    & \ctrl_i[t]\in \Ubox,~ i = 1, \ldots, N \label{eqn: box_constraint}
\end{align}
\end{subequations}


where $\nabla h$ and $\nabla V$ are the generalized gradients of the CBF and CLF respectively. Note that \eqref{eq:central_safe} has the form of a CLF-CBF controller defined in \cite{ames_cbf_2019}.
Constraint \eqref{eqn: online_problem_cbf} is the \textit{safety} constraint which uses $h$ with coefficient $\alpha>0$ to ensure all robots avoid obstacles and remain mutually separated.
Constraint \eqref{eqn: online_problem_clf} uses $V$ with $\lambda>0$ to ensure the system is always driven towards the goal region. 
The slack variable $\delta$ in Constraint \eqref{eqn: online_problem_clf} is unbounded to allow relaxation of this constraint, as it is lower priority to \eqref{eqn: online_problem_cbf}.
We solve \eqref{eq:central_safe} centrally following the approach in \cite{magnus_2017_NBF} for quadratic-program based controllers with non-smooth barrier functions.

Alternatively, to solve the safe control problem distributedly, we adapt the approach from \cite{do_margellos_2023a}. 
Each robot solves:

\begin{subequations} \label{eqn:dist_safe}
\begin{align} 
    \min_{\ctrl_k[t]} ~&~ J_k(\ctrl_k[t], \sigma(\ctrl[t])) \label{eqn: dist_problem_cost} \\
    \text{s.t. } ~&~ \mathbf{C}_k[t] \ctrl_k[t] + \sum_{j=1, j\neq k}^N \mathbf{C}_j[t] \ctrl_j[t] \leq \sum_k^N b_j[t] \label{eqn: dist_problem_constraints}
\end{align}
\end{subequations}

Matrices $\mathbf{C}_k[t]$ and vectors $b_k[t]$ in \eqref{eqn: dist_problem_constraints} incorporate the CBF and CLF constraints for localized for each robot $k$. We refer to \cite{do_margellos_2023a} for how these are constructed. The cost function

\begin{subequations}
\begin{align}
    J_k(\ctrl_k[t], \sigma(\ctrl[t])) & = \| \ctrl_k[t] - \ctrl_k^{fair}[t] \|^2 + \delta_k^2 + \sigma(\ctrl[t])
\end{align}
\end{subequations}

penalizes the distance of the inputs of robot $k$ from their fair control inputs. The term $\sigma(\ctrl[t]) = \frac{1}{N} \sum_{k} \| \ctrl_k[t] - \ctrl_k^{fair}[t] \|^2$ drives minimization of the team's total distance from the fair control inputs, and the slack variable $\delta_k$ allows relaxation of robot $k$'s localized CLF constraint.

\begin{algorithm}
\caption{FiReFly}
\label{alg:full_sol}
\begin{algorithmic}[1]
\State{Input: max time $H$, $\ctrl^{(0)}$, $\state[0]$, $\mathbf{A}_k, \mathbf{B}_k, ~1 \leq k \leq N$}
\State{Initialize $\ctrl^{\leq t}$ as empty}
\For{$t = 1, \ldots, H$}
    \State{\textbf{do} Algorithm 1, obtain $\ctrl^{fair}$}
    \State{\textbf{solve} \eqref{eq:central_safe} or \eqref{eqn:dist_safe}, obtain $\ctrl^{safe}[t]$}
    \State{update $\state_k[t+1] = \mathbf{A}_k\state_k[t-1] + \mathbf{B}_k\ctrl^{safe}_k[t], \forall k$}
    \State{$\ctrl^{\leq t}[t] \leftarrow \ctrl^{safe}_k[t]$}
\EndFor
\end{algorithmic}
\end{algorithm}

Algorithm \ref{alg:full_sol} describes our full solution for FiReFly. Given system dynamics and initial state for all robots, at each timestep $t$, first do Algorithm \ref{alg:distributed_opt} to obtain $\ctrl^{fair}$, the length $H$ fair sequence of inputs for the team. Then, obtain the safe inputs $\ctrl^{safe}[t]$ using either by solving centrally with \eqref{eq:central_safe} or distributedly with each robot solving \eqref{eqn:dist_safe}. Drive the system forward with these inputs, and append them to the prefix. Then repeat, solving for the next $H-1$ input sequences. 

%% file: sections/experiments.tex
We conduct two types of experiments to evaluate FiReFly under different fairness notions. The first investigates how \ourmethod~performs with increasing number of obstacles. The second investigates scaling of the number of UAVs.
For all experiments, we set $\Ubox = [-100, 100]$, with safe mutual separation distance $d_s = 0.01$, planning horizon $H=25$, and sample time $d_t = 0.2$. In fairness notions $f_2$ and $f_4$, we use $\beta=10^{-6}$ to scale the objective's components. In $f_3$ and $f_4$ we set $M=10$. 
For the local problem \eqref{eq:dist_fair}, we set $\varepsilon_{min/max} = \pm 10$, and $\kappa=1$. The initialized control inputs $\ctrl^{(0)}$ to Algorithm \ref{alg:full_sol} produce a simple straight-line trajectory as computed by the minimum-jerk motion planner from \cite{andrea_2015}. 
We set the maximum iterations for Algorithm \ref{alg:distributed_opt} to $1000$ and use a stopping criterion of $\|\ctrl^{(r)} - \ctrl^{(r-1)}\| \leq \eta$ where $\eta=0.5$ if using $f_1$ or $f_2$, and $\eta=0.1$ if using $f_3$ or $f_4$. For the CBF and CLF constraints, we set $\alpha=0.15$ and $\gamma=0.025$ in the central formulation and $\alpha=0.1$ and $\gamma=0.1$ in the distributed formulation. 
Simulations were implemented in Python, with \eqref{eq:dist_fair} and \eqref{eqn:dist_safe} solved using JAXopt \cite{jaxopt_implicit_diff} and \eqref{eq:central_safe} solved with MOSEK \cite{mosek}. All experiments ran on a partition of a high-performance computer cluster using 24 cores with a 2.1 Ghz processor and with access to 50GB RAM. 

\subsection{Experiment 1: Increasing Number of Obstacles}



In this experiment we fix the number of UAVs ($N=5$) and their starting positions.
Obstacles are uniformly randomly generated somewhere along the straight-line path between a UAV's starting position and a single goal area. 
We generate 200 trial configurations with the number of obstacles varying from 1 to 5. For each configuration, we run FiReFly for all fairness notions with both the central \eqref{eq:central_safe} and distributed \eqref{eqn:dist_safe} safe control formulations. We also run FiReFly with central safe control and without any fairness notions as a baseline. 

\begin{figure}[htp]
\centering
{\includegraphics[width=.79\linewidth]{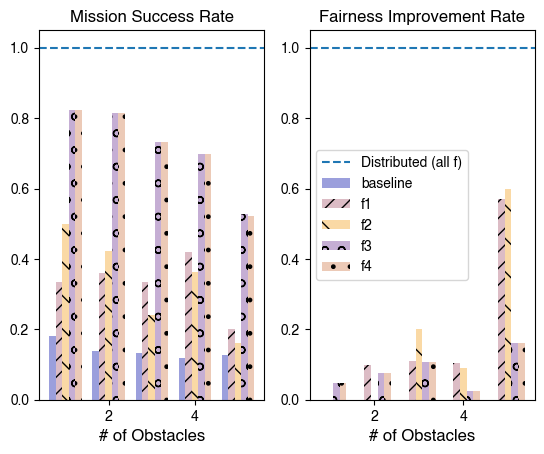}}
\caption{Results for Experiment 1: 5 UAVs with increasing number of obstacles. Mission success rates (left) and the rate of fairness improvement over the baseline (right) for \ourmethod~with central safe controller under different notions of fairness (bars). \ourmethod~with distributed safe controller yielded 100\% mission success rate and fairness improvement rate for all fairness notions (horizontal line). The legend on the right chart also applies to the left.}
\label{fig:exp1_results}
\end{figure}

\textbf{Mission Success:} We define the mission success rate as the total number of UAVs to reach their goal area (within $H$ time units) divided by the total number of UAVs in all trials. FiReFly with distributed safe control achieves better 100\% mission success rates as shown in Fig. \ref{fig:exp1_results}. FiReFly with central safe control also improves mission success over the baseline, though we see this success rate decrease as the number of obstacles increases.

\textbf{Fairness:} FiReFly with distributed safe control also improves fairness over the baseline in 100\% trials. In contrast, FiReFly with central safe control improved fairness over the baseline only a small fraction of the time.

\begin{table*}[htp]
\vspace*{0.12cm}
\centering
\resizebox{.7\textwidth}{!}{%
\begin{tabular}{l|lll|llllll|}
\cline{2-10}
 & \multicolumn{3}{c|}{FairFly \cite{kurtz20fairfly}} & \multicolumn{6}{c|}{Distributed FiReFly (this work)} \\ \cline{2-10} 
 & \multicolumn{3}{c|}{Total Runtime} & \multicolumn{3}{c|}{Fair Planner Runtime} & \multicolumn{3}{c|}{Safe Control Runtime} \\ \hline
\multicolumn{1}{|c|}{\# of UAVs} & \multicolumn{1}{c|}{mean} & \multicolumn{1}{c|}{std} & \multicolumn{1}{c|}{max} & \multicolumn{1}{c|}{mean} & \multicolumn{1}{c|}{std} & \multicolumn{1}{c|}{max} & \multicolumn{1}{c|}{mean} & \multicolumn{1}{c|}{std} & \multicolumn{1}{c|}{max} \\ \hline
\multicolumn{1}{|l|}{7} & \multicolumn{1}{l|}{35.742} & \multicolumn{1}{l|}{12.132} & 46.462 & \multicolumn{1}{l|}{\textbf{0.005}} & \multicolumn{1}{l|}{0.007} & \multicolumn{1}{l|}{0.031} & \multicolumn{1}{l|}{0.011} & \multicolumn{1}{l|}{0.000} & 0.012 \\
\multicolumn{1}{|l|}{10} & \multicolumn{1}{l|}{78.644} & \multicolumn{1}{l|}{17.453} & 104.051 & \multicolumn{1}{l|}{\textbf{0.005}} & \multicolumn{1}{l|}{0.006} & \multicolumn{1}{l|}{0.029} & \multicolumn{1}{l|}{0.014} & \multicolumn{1}{l|}{0.001} & 0.015 \\
\multicolumn{1}{|l|}{12} & \multicolumn{1}{l|}{63.446} & \multicolumn{1}{l|}{29.451} & 102.503 & \multicolumn{1}{l|}{\textbf{0.007}} & \multicolumn{1}{l|}{0.007} & \multicolumn{1}{l|}{0.031} & \multicolumn{1}{l|}{0.019} & \multicolumn{1}{l|}{0.001} & 0.019 \\
\multicolumn{1}{|l|}{15} & \multicolumn{1}{l|}{109.711} & \multicolumn{1}{l|}{23.950} & 138.048 & \multicolumn{1}{l|}{\textbf{0.010}} & \multicolumn{1}{l|}{0.008} & \multicolumn{1}{l|}{0.033} & \multicolumn{1}{l|}{0.021} & \multicolumn{1}{l|}{0.001} & 0.024 \\
\multicolumn{1}{|l|}{20} & \multicolumn{1}{l|}{TIMEOUT} & \multicolumn{1}{l|}{TIMEOUT} & TIMEOUT & \multicolumn{1}{l|}{\textbf{0.007}} & \multicolumn{1}{l|}{0.009} & \multicolumn{1}{l|}{0.029} & \multicolumn{1}{l|}{0.024} & \multicolumn{1}{l|}{0.001} & 0.025 \\
\multicolumn{1}{|l|}{50} & \multicolumn{1}{l|}{TIMEOUT} & \multicolumn{1}{l|}{TIMEOUT} & TIMEOUT & \multicolumn{1}{l|}{\textbf{0.014}} & \multicolumn{1}{l|}{0.002} & \multicolumn{1}{l|}{0.016} & \multicolumn{1}{l|}{0.053} & \multicolumn{1}{l|}{0.001} & 0.054 \\ \hline
\end{tabular}%
}
\caption{Runtimes in seconds for FairFly and FiReFly. For FiReFly, runtimes for the fair motion planner and the safe controller are split. Timeout is declared after 5 minutes. Both \ourmethod~steps take considerably less time than FairFly, with fair planning being the quickest.}
\label{tab:runtimes}
\end{table*}

\begin{figure}[htp]
\centering
{\includegraphics[width=.79\linewidth]{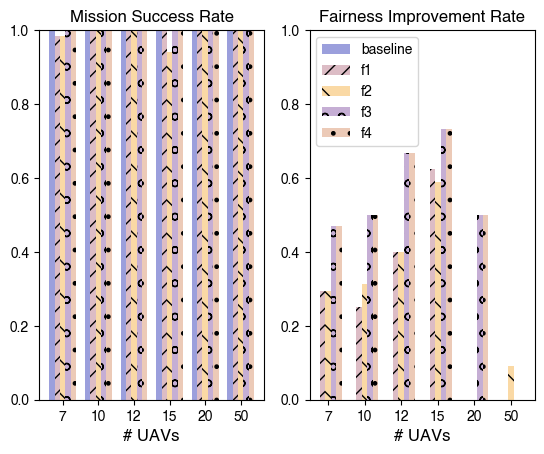}}
\caption{Experiment 2: Mission success rates (left) and Fairness improvement rates (right) vs number of UAVs for FiReFly with distributed safe control. Nearly 100\% mission success is achieved for all fairness notions. Fairness improvements increase with the UAV team size up to 15.}
\label{fig:exp2}
\end{figure}

\subsection{Experiment 2: Scaling number of UAVs}
Scaling the UAV team size increases the number of iterations for convergence in Algorithm 1, and the safety constraints. In this experiment, we assess these scaling effects on runtime of \ourmethod~for UAV team sizes of $N\in \{7,10,12,15,20,50\}$. 
We fix a single obstacle at with radius and randomly place UAVs starting and goal positions around it. 
For each team size, we generate 20 of these random configurations and run \ourmethod~for all fairness notions. 
We focus on distributed \ourmethod~in this experiment given that it had the better performance over the two different safe control formulations in Experiment 1. Mission success rates were nearly 100\% for all team sizes.

\textbf{Runtime:} We compare the runtime of \ourmethod~to that of FairFly \cite{kurtz20fairfly}, a state of the art UAV motion planner, in Table \ref{tab:runtimes}. For both approaches, the reported runtimes are for the time it takes to compute an input for a single timestep. FiReFly is faster than FairFly in all UAV team sizes, and growth in runtime is also slower.

\textbf{Fairness:} Fairness generally improved as team size grew. We see in this in Fig \ref{fig:exp2} with the exception of 20 and 50 UAVs. In these cases, fairness degrades to the traffic density necessitating more control adjustments for safety. To increase fairness improvements for the larger UAV teams, we tried to tighten the convergence criterion in Algorithm \ref{alg:distributed_opt} (smaller $\eta$). In this case, fairness improved nearly 70\% of the time for 50 UAVs, but runtime increased by a factor of $10^3$.
To reduce runtime for 50 UAVs, we ran FiReFly without any online fair re-planning (i.e. skipping line 4 in Algorithm~\ref{alg:full_sol}), simply using the safe controller to track the initially planned fair trajectory. We found that while this saves computation time, fairness worsens.
Thus, a balance can be achieved by re-planning every few steps with a tight convergence criterion.

%% file: sections/appendix.tex
\begin{lemma}
\label{lemma:local problem}
The function 
$h(\varepsilon) = -\varepsilon^T \nabla_{\ctrl_k}~ f(\ctrl^{(r-1)}) + \kappa \|\varepsilon\|^2$
is continuously differentiable in $\varepsilon$, and strongly convex.
Moreover, $\nabla_{\ctrl_k}f(\ctrl)$ is Lipschitz continuous on $\Ubox$.    
\end{lemma}

\textit{Proof of Lemma 1}:   
Function $h$ is convex being the sum of a linear term $\varepsilon^T \nabla_{\ctrl_k}~ f(\ctrl^{(r-1)})$ (the gradient being constant in $\varepsilon$) and a quadratic term $\|\varepsilon\|^2$. 
We can choose $\kappa > 0$ large enough such that $h$ also strongly convex. A function $g(x)$ is strongly convex if for $c > 0$, $g(x) - \frac{c}{2}\|x\|^2$ is convex (see \cite{Boyd_Vandenberghe_2004} Chp. 9). 
Function $h$ is also clearly continuously differentiable. 
Finally, all fairness functions $f$ are twice differentiable, and therefore the gradient is Lipschitz continuous.

\textit{Proof of convergence of Algorithm \ref{alg:distributed_opt}}:
The update step of Algorithm \ref{alg:distributed_opt} is $\ctrl_k^{(r)}=\ctrl_k^{(r-1)} + \gamma^{(r)} \varepsilon_k^{(r)}$ where we set $\gamma^{(r)}$ to gradually decrease from 1 to 0.1 every iteration. This choice of $\gamma^{(r)}$ meets the following conditions: 1) $\gamma^{(r)} \in (0, 1]$, 2) $\sum_{r=0}^{\infty} \gamma^{(r)} = +\infty$, and 3) $\lim \sup_{r \rightarrow \infty} \gamma^{(r)} < C$, where $C$ is a positive constant. 
Given this, that the team inputs are updated in a cyclic manner (each robot $k$ is picked on after the other), and Lemma \ref{lemma:local problem}, it follows from the convergence properties in Theorem 1 of \cite{razaviyan14dstr} that Algorithm \ref{alg:distributed_opt} converges to a stationary point of $f$.